\documentclass[10pt,twocolumn,letterpaper]{article}

\usepackage[pagenumbers]{cvpr} 

\usepackage{graphicx}
\usepackage{amsmath}
\usepackage{amssymb}
\usepackage{booktabs,arydshln}
\usepackage{microtype}
\usepackage{algorithm,algpseudocode}
\usepackage[dvipsnames,x11names]{xcolor,colortbl}
\usepackage{xparse}
\usepackage{makecell}
\usepackage{comment}
\usepackage{multirow}

\usepackage[pagebackref,breaklinks,colorlinks]{hyperref}

\usepackage[capitalize]{cleveref}
\crefname{section}{Sec.}{Secs.}
\Crefname{section}{Section}{Sections}
\Crefname{table}{Table}{Tables}
\crefname{table}{Tab.}{Tabs.}

\def\cvprPaperID{21} 
\def\confName{CVPR}
\def\confYear{2023}

\begin{document}

\title{Continual Learning for LiDAR Semantic Segmentation:\\Class-Incremental and Coarse-to-Fine strategies on Sparse Data}

\author{Elena Camuffo and Simone Milani\\
Department of Information Engineering, University of Padova, Italy\\
{\tt\small \{elena.camuffo,simone.milani\}@dei.unipd.it}
}
\maketitle

\begin{abstract}
{During the last few years, Continual Learning (CL) strategies for image classification and segmentation have been widely investigated designing innovative solutions to tackle catastrophic forgetting, like knowledge distillation and self-inpainting.}
However, the application of continual learning paradigms to point clouds is still unexplored and investigation is required, especially using architectures that capture the sparsity and uneven distribution of LiDAR data.
The current paper analyzes the problem of class incremental learning applied to point cloud semantic segmentation, comparing approaches and state-of-the-art architectures. To the best of our knowledge, this is the first example of class-incremental continual learning for LiDAR point cloud semantic segmentation.
Different CL strategies were adapted to LiDAR point clouds and tested, tackling both classic fine-tuning scenarios and the Coarse-to-Fine learning paradigm. 
The framework has been evaluated through two different architectures on SemanticKITTI \cite{behley2019iccv,geiger2012cvpr}, obtaining results in line with state-of-the-art CL strategies and standard offline learning.
\end{abstract}

\section{Introduction}\label{sec:intro}
The recent development of autonomous driving, equipped with automatic visual understanding systems, 
{has fostered the research on object identification and segmentation towards more generalizable and portable strategies via inline learning behaviors in place of distinct offline training phases.}
Particularly, continual learning investigates the capability of incorporating new knowledge in well-established models, shifting to new tasks, labels, or data distributions.  Indeed, most of the cases do not benefit from previous training data, which are unavailable due to portability, privacy issues, or costly labeling process. The process is thus usually accomplished by preserving previous knowledge, whilst avoiding catastrophic forgetting.
Different CL strategies have been deeply investigated for classification \cite{9349197,kirkpatrick_overcoming_2017,Li2016LearningWF,icarl} and object detection \cite{8237630,joseph2021towards,Peng2020faster} tasks, while less examples exist for semantic segmentation. The reason is that semantic segmentation is a harder task with respect to classification: {classification is about providing one single class label for each image, while in segmentation already-known and totally-new classes can be present together in the same acquisition (image, point cloud, etc.).}
{As a matter of fact, handling with such heterogeneity is one} of the additional issues that is addressed considering multiple continual learning scenarios for semantic segmentation \cite{michieli2019,klingner2020class,douillard2021plop,cermelli2022incremental,toldo2022learning}.
{So far, continual learning has generally} been applied to images in vision tasks. {With respect to point cloud data,} images are far simpler data to be retrieved, labeled, and processed {thanks to their regular distribution in space, limited memory requirements, and computational cost.}
Instead, {CL} techniques on point cloud data are still {quite unexplored}, made an exception for a few attempts \cite{cen2022open,LIN202073,Dong2020I3DOLI3,knights2022incloud,Chowdhury2022Few}, {and the investigation of class-incremental CL} has not been considered yet. Even if {it may} seem a simple adaptation of image-based network architectures to point cloud, {the sparsity of LiDAR point cloud data} and the significant difference in the {processing architectures makes their investigation anew}. {More precisely, the problem turns out to be more challenging in relation to point-based architectures \cite{qi2016pointnet,qi2017pointnetplusplus,hu2019randla}, where raw points are directly processed with no discretization (see Sec.~\ref{sec:related} for more details). In addition, the inherent nature of point cloud datasets, acquired with different methodologies and in different noise conditions, can vary the final accuracy.}

{This paper analyzes the challenges of class-incremental continual learning on point cloud semantic segmentation  and adapts different CL solutions to different architectures and partitionings.} 
{The main novelties brought by this work can be listed as follows.
\begin{itemize}
\item{It is the first application of class-incremental continual learning to LiDAR semantic segmentation, where point sparsity enhances the issues concerning the co-occurrence of already-known and totally-new classes.}
\item{It provides a general evaluation of  different techniques to mitigate catastrophic forgetting on LiDAR point clouds (adapted from image semantic segmentation to a domain with new peculiarities).}
\item{It considers both general Class-Incremental and Coarse-to-Fine scenarios \cite{li2021prototypical,yang2021part,mel2020incremental,stretcu2020coarse}.}
\end{itemize}
}

{The paper is structured as follows. Section \ref{sec:related} overviews recent state-of-the-art works, 
Section \ref{sec:problem} presents the main single-headed class-incremental continual learning scenarios. Section \ref{sec:method} analyzes different CL solutions overviewing different partitioning and strategies to mitigate catastrophic forgetting.
Finally, in Section \ref{sec:exp} we discuss the results obtained with different techniques with a comparison among dataset setups and CL strategies; conclusions are drawn in Section \ref{sec:conclusion}.}

\section{Related Work}\label{sec:related}

\textbf{LiDAR Semantic Segmentation} is usually tackled via deep learning solutions that treat input points in several ways: discretizing into voxels, projecting on 2D images, or processing directly as they are \cite{camuffo2022recent,gao2021we}. 
Voxel based methods \cite{Tchapmi2017SEGCloudSS,7353481} are generally used for well-structured point clouds (\eg, terrestrial laser scanning acquisitions \cite{armeni20163d}) and can easily apply 3D convolutions. Projection based methods \cite{milioto2019rangenet++,zhang2020polarnet,cortinhal2020salsanext} are well suited for LiDAR point clouds, as they are acquired in concentric circles and can be easily projected onto cylindrical \cite{zhang2020polarnet} or spherical \cite{milioto2019rangenet++} surfaces. Both categories of methods are efficient in terms of computational complexity, but approximate point distributions to regular structures.
Methods that directly process data \cite{qi2016pointnet,qi2017pointnetplusplus,hu2019randla} avoid such loss of structural information, which is crucial when dealing with few data samples and limited resources. Recent works rely on 4D convolutions \cite{fan2021pstnet} or transformers \cite{Guo_2021} to accomplish the task but require huge computational power and storage capacity; more lightweight approaches process points in mixed strategies, providing voxel-wise predictions refined with point-wise labels \cite{zhu2021cylindrical,zhu2021cylindrical-tpami}.
Such models have been widely adopted for various learning approaches including class-balancing regularization \cite{Camuffo2023,cortinhal2020salsanext} Knowledge distillation \cite{Hou_2022_CVPR} and open-world 3D scene understanding \cite{cen2022open}. Some works have also adopted incremental learning settings for 3D data in object detection \cite{Dong2020I3DOLI3}, place recognition \cite{knights2022incloud}, remote sensing \cite{LIN202073} and few-shot learning \cite{Chowdhury2022Few}.
Nonetheless, no prior work has explored Class-Incremental Continual Learning for LiDAR Semantic Segmentation.

\textbf{Class-Incremental Continual Learning (CL)} has developed a growing research interest for image classification \cite{9349197,kirkpatrick_overcoming_2017,Li2016LearningWF,icarl}, object detection \cite{8237630,joseph2021towards,Peng2020faster} tasks. Many of these works observed the catastrophic interference problem when sequentially learning examples of different input patterns. Among the most popular strategies to reduce forgetting are replay methods \cite{icarl,maracani2021recall} that store exemplars to be replayed in future steps and non-exemplar methods; among these 
 parameter isolation dedicates different model parameters to each task to prevent any possible forgetting and regularization methods \cite{Li2016LearningWF,kirkpatrick_overcoming_2017,toldo2022learning,cermelli2020modeling}, which add extra regularization loss to consolidate knowledge on previous data. This line of work avoids storing raw inputs, prioritizing privacy, and alleviating memory requirements.

\textbf{CL in Semantic Segmentation} literature is more limited \cite{michieli2019,klingner2020class,douillard2021plop,cermelli2022incremental,toldo2022learning}. Major attention has been devoted to class incremental semantic segmentation to learn new categories from new data. The problem is formalized in \cite{michieli2019} and tackled with regularization approaches such as parameters freezing and knowledge distillation. In \cite{klingner2020class} knowledge distillation is coupled with a class importance weighting scheme to emphasize gradients on difficult classes. In \cite{michieli2021continual} the latent space is regularized to improve class-conditional features separation. Cermelli \etal study the distribution shift of the background class in \cite{cermelli2020modeling} and tackle the problem in a weakly-supervised fashion in \cite{cermelli2022incremental}. 
Maracani \etal \cite{maracani2021recall} proposed replaying old classes using either generative networks or web crawled images.

\textbf{Coarse-to-Fine Semantic Segmentation} has been explored in few previous work for part-based regularization \cite{li2021prototypical,yang2021part} and approaches where coarse-level classes are refined into finer categories \cite{mel2020incremental,stretcu2020coarse}. 
Previous approaches developed also different ways to draw class hierarchies. In \cite{shenaj2022continual} they are estimated a prior based on semantics; in \cite{Camuffo2023}, they are drawn a posteriori based on network misclassifications.
Some examples have shown also developing Coarse-to-Fine for continual learning \cite{shenaj2022continual,xiang2022coarse}, but no previous literature has explored with point cloud data.

\begin{table*}[t]
    \centering
    \setlength{\tabcolsep}{0.5em}
    \renewcommand{\arraystretch}{0.9} \small
    \begin{tabular}{ccccc}
    \toprule
        & Training Subset $T_0$ & Training Subset $T_1$ & Training Subset $T_2$ & Validation Set \\ \midrule
        & \multicolumn{3}{c}{SemanticKITTI train} & SemanticKITTI val \\ \midrule
         Sequences & $D_0=$\texttt{\{01,02,03\}} & $D_1=$\texttt{\{04,05,09,10\}} & $D_2=$\texttt{\{00,06,07\}}& $D=$\texttt{\{08\}} \\ 
         \makecell{Labeled\\classes} & \it \makecell{ $C_0 =$ \{road, \underline{parking},\\sidewalk, \underline{other-ground},\\vegetation, terrain\}} & \it \makecell{ $C_1 =$ \{building, fence,\\\underline{trunk}, pole, traffic-sign\}}  & \it \makecell{ $C_2 =$ \{bicycle, motorcycle,\\truck, \underline{other-vehicle}, person,\\ \underline{bicyclist}, \underline{motorcyclist}, car\}} & \it \makecell{ $C = \{C_0 \cup C_1 \cup C_2\}$}\\ 
         & \begin{minipage}{.2\textwidth}
         \includegraphics[trim=0cm 0cm 0cm 0cm, clip, width=\textwidth]{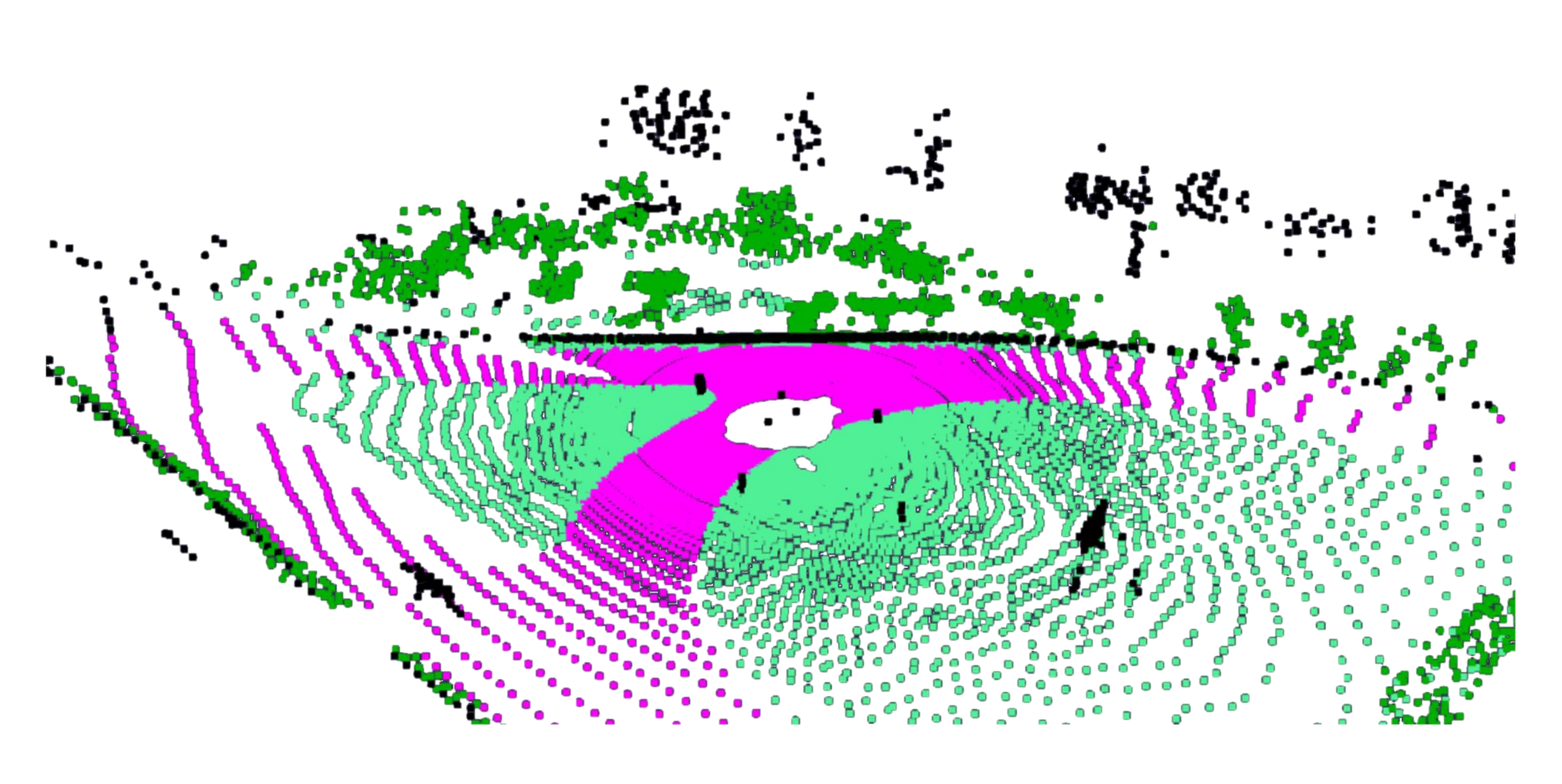}
         \end{minipage} 
         & \begin{minipage}{.2\textwidth}
         \includegraphics[trim=0cm 0cm 0cm 0cm, clip,width=\textwidth]{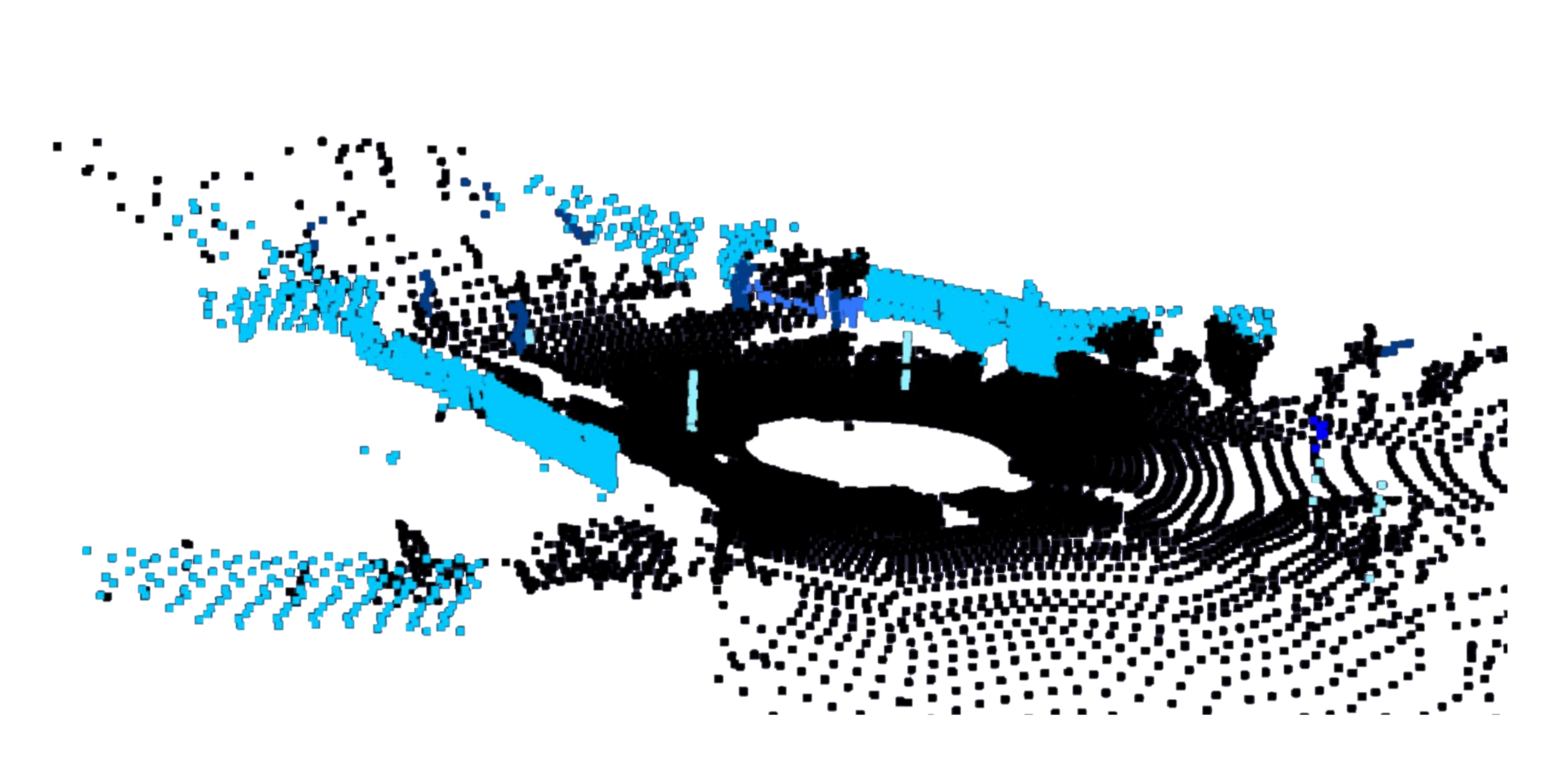} 
         \end{minipage} &
         \begin{minipage}{.2\textwidth}
         \includegraphics[trim=0cm 0cm 0cm 0cm, clip,width=\textwidth]{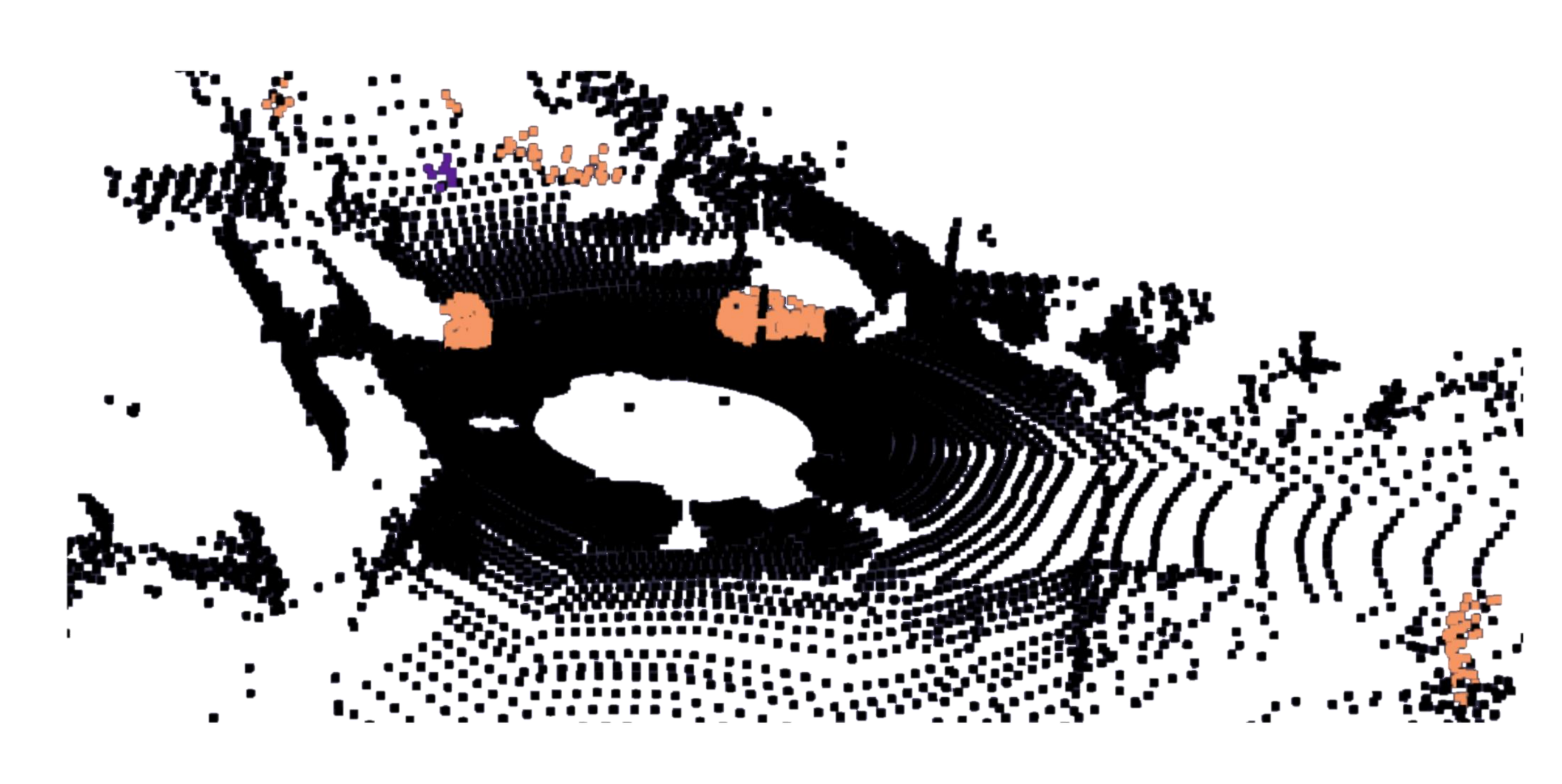} \end{minipage}         &
         \begin{minipage}{.2\textwidth}
         \includegraphics[trim=2cm 2cm 0cm 0cm, clip,width=\textwidth]{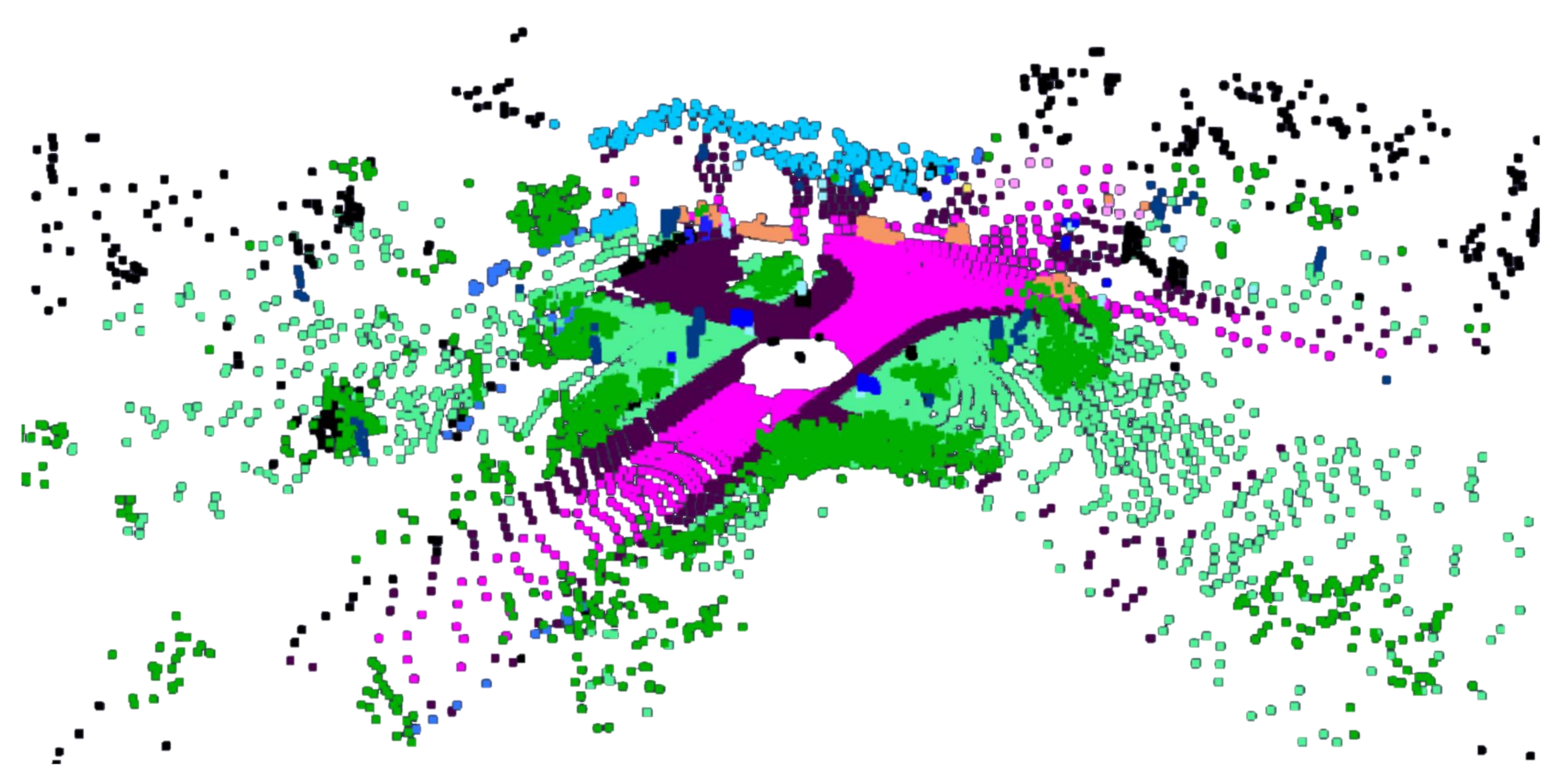}\end{minipage} \\
         \# Clouds & $6563$ & $4623$ & $4541$ & $4071$\\
         \# Points & $355280$ & $375140$ & $121494$ & $19130$ \\
         \makecell{\% Labeled pt.} & $81.73$\% & $19.21$\% & $8.82$\% & $95.53\%$\\
         \bottomrule
    \end{tabular}
    \caption{SemanticKITTI \cite{behley2019iccv} subdivision in experiences. Training set classes and sequences are divided in three parts according to points frequency and cardinality, whilst following \cite{klingner2020class}. \underline{Underline} denotes classes that are not present in Cityscapes \cite{Cordts_2016_CVPR}, thus in the original partition.}
    \label{tab:dset-subdivision}
\end{table*}

\begin{figure*}
    \centering 
    \includegraphics[width=\textwidth]{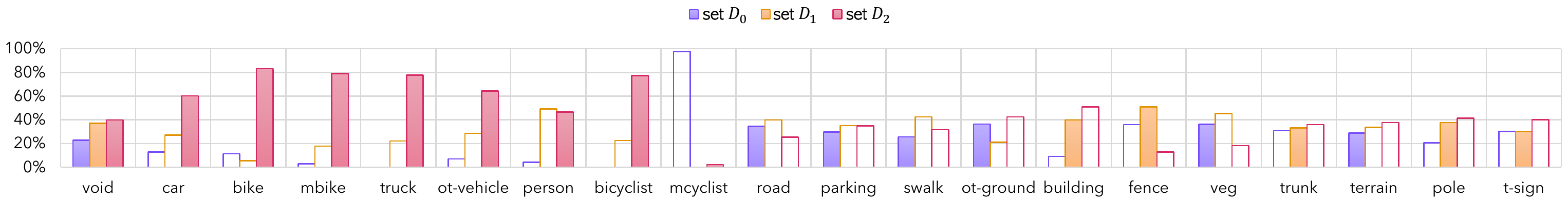}
    \caption{Histogram of point clouds percentage within each training dataset split $D_k$. The percentage is normalized for each class $c$ on the total number of points for that specific class in the whole dataset $D_0 \cup D_1 \cup D_2$.}
    \label{fig:chart}
\end{figure*}

\section{Problem Formulation}\label{sec:problem}

Given an input point cloud, \ie, a set of $N$ 3D points ${\bf X} = \{ {\bf x}_1, {\bf x}_2, \dots , {\bf x}_N \} $, and a set of candidate semantic labels $Y = \it \{y_1, y_2, \dots , y_N \}$, the objective of semantic segmentation is to associate each input point ${\bf x}_i$ with a semantic label $y_i$. Each label belongs to a given subset of semantic classes $C$ that contains also a special class $b$ (\textit{void}, which corresponds to the \textit{background} for images). {The void class typically identifies} objects out of the range of LiDAR precision or belonging to unknown classes. {For the sake of clarity}, we call it \textit{background} as for images to avoid confusion with empty voxels and masked points.
Semantic segmentation task is usually achieved by using a suitable deep learning model $\mathcal{M}: X \mapsto \hat{Y}_{C}$, commonly composed of a feature extractor $\mathcal{E}$ followed by a decoding module $\mathcal{D}$, $\mathcal{M} = \mathcal{D} \circ \mathcal{E}$.

In standard supervised learning, the model is learned in a single shot with {standard offline training over a fixed and complete set of data $T \subset X \times Y$}. 

In class incremental learning, instead, we assume that the training is performed in multiple steps {and only a subset of the training data is available for training} at each step $k = 0, \dots, K$.
More in detail, we start from an initial step $k = 0$ where only training data concerning a subset of all the classes $C_0 \subset C$ is available (we assume that $b \in C_0$).
We denote with $\mathcal{M}_0 : X \mapsto \hat{Y}_{C_{0}}$, $\mathcal{M}_0 = \mathcal{D}_0 \circ \mathcal{E}_0$ the model trained after this initial step.
{At a generic step $k$}, a new set of classes $C_k$ is added to the class collection $C_{0\rightarrow(k-1)}$ learned up to that point, resulting in an expanded set of learnable classes $C_{0\rightarrow k} =C_{0\rightarrow(k-1)} \cup C_k$ (we assume $C_{0\rightarrow(k-1)} \cap C_k = \emptyset$). The model after the $k$-th step of training is $\mathcal{M}_k : X \mapsto \hat{Y}_{C_{0\rightarrow k}}$, where $\mathcal{M}_k = \mathcal{D}_k \circ \mathcal{E}_k$.\\
 Some approaches train the decoder alone after the initial step while keeping the encoder frozen. In our approach, both the encoder and decoder are updated, while only the last layer is changed.
This is because the encoder is usually kept frozen when the training is performed on random objects datasets, and the proportion {of data} between the first step and the subsequent ones is unbalanced toward the first. {Instead, in our {partitioning} (and in general, in autonomous driving datasets), dataset cardinality, number of samples, and frequency of classes are balanced across the different steps.}

\subsection{Scenarios} \label{ssec:scenarios}

{Multiple scenarios have been proposed for CL on image semantic segmentation\cite{MICHIELI2022275,michieli2019}.
In each scenario, the original dataset $T$ is divided into $K$ groups that correspond to learning steps; the overall set of labels is also generally divided into $K$ groups.}
{At learning step $k$, the training procedure updates the model using only the tuple $(T_k, C_k)$.}

\textbf{Sequential}: samples in the $k$-th experience are provided with the labels of all the groups $\leq k$, while points with labels $>k$ are not present at all. Each learning step contains a unique set of samples, whose points belong to classes seen either in the current or in the previous steps. 

\textbf{Sequential masked}: samples in the $k$-th experience are provided with the $k$-th experience labels, while the points with labels of classes $<k$ are marked as \textit{unlabeled}, and points with labels $>k$ are not present at all.
In this setup, each learning step contains a unique set of samples, whose points belong either to novel classes or to the \textit{unlabeled} class, which is not predicted by the model and is masked out from both the results and the training procedure.

\textbf{Disjoint}: samples in the $k$-th experience are provided with the $k$-th experience labels, while points with labels of classes $<k$ are marked as \textit{background}, and points with labels $>k$ are not present at all. At each learning step, the unique set of training samples is identical to the sequential setup, {but old classes are gathered into a common label (\textit{background}, instead of being unlabelled as in the Sequential masked scenario) that changes its distribution at each step.}

\textbf{Overlapped}: At each learning step $k$ all the dataset samples are available but only the current step classes $C_k$ are labeled with the correct label and the rest ($C_j \neq C_i$) with the background labels, either if they belong to experience $j<k$ or $j>k$. In this setup, each training step contains all the point clouds that have at least one point belonging to a class in set $C_k$, with only the classes of the set annotated and the rest set to \textit{background}.
{In this paper, this scenario is reported for comparison assuming that data can never be shared among different experiences.}

\textbf{Coarse-to-Fine}: At each learning step $k$, {all points are labeled but their classes are refined with respect to the previous step.} At step $k$, {coarse labels $C_{k-1}$ are substituted with labels from fine classes $C_k, |C_k| > |C_{k-1}|$. Each coarse class corresponds to its unique set of fine classes. 

\begin{figure}[t]
    \centering
    \includegraphics[width=\linewidth]{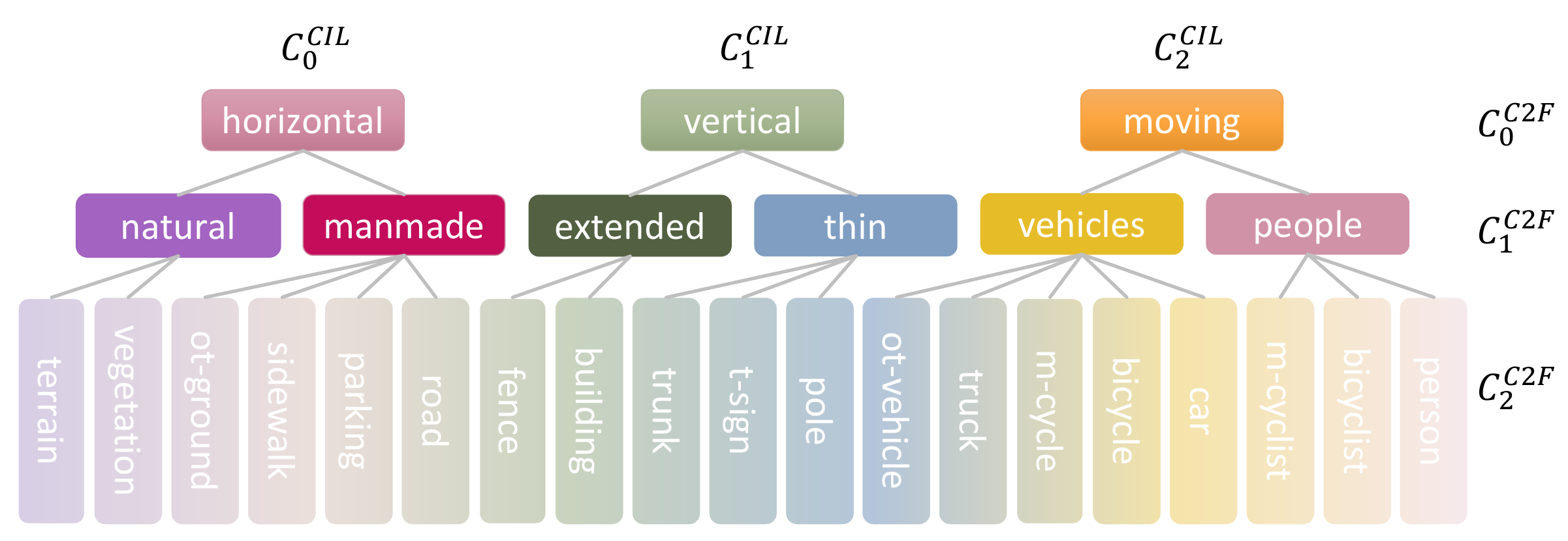}
    \caption{Class hierarchies in SemanticKITTI \textbf{C2F} partition. Each Coarse class is made-up by grouping classes belonging to the same learning step in \textbf{CIL} partition. Sequence splits $D_k$ are considered in the same identical order.}
    \label{fig:hierarchy}
\end{figure}

\section{Methodology}\label{sec:method}
We considered \textbf{RandLA-Net} \cite{hu2019randla}, one of the most famous point-based architectures as a reference, in order to frame the problem in a perspective different from images (\ie, using MLPs in place of convolutional setting). {RandLA-Net} is an efficient point-based lightweight network composed of an MLP based encoder-decoder structure that achieves remarkably high efficiency in terms of memory and computation. In addition, we evaluate on \textbf{Cylinder3D} \cite{zhu2021cylindrical,zhu2021cylindrical-tpami} voxel based architecture for comparison. Nonetheless, the framework can be applied on top of any architecture for point cloud semantic segmentation.

\textbf{SemanticKITTI} \cite{behley2019iccv,geiger2012cvpr} has been chosen as a reference dataset, since it is one of the most popular benchmarks for LiDAR semantic segmentation in autonomous driving. SemanticKITTI consists of $43,552$ densely annotated LiDAR scans, $19,130$ for training, and $4,071$ for validating (that we used for testing, as done by all competing works being the test labels not publicly available). 

\subsection{Dataset Partitioning}\label{sec:dataset}

We divided the dataset in order to define incremental experiences for continual learning. As no previous works can be found, we propose two subdivisions of SemanticKITTI: one based on class cardinality and frequency within different sequential acquisitions, and another based on a Coarse-to-Fine semantic partitioning. We denote the former as \textbf{CIL}, as it is based on the effective subdivision of \cite{klingner2020class}, and the latter as \textbf{C2F}.

\subsubsection{Semantic Partition (CIL)}\label{ssec:cil}
 Tab.~\ref{tab:dset-subdivision} shows the first partition of SemanticKITTI, based on class cardinality and frequency within different sequential acquisitions. {This grouping is based on the effective subdivision of Cityscapes \cite{Cordts_2016_CVPR} proposed in \cite{klingner2020class}. Cityscapes is an autonomous driving dataset for image semantic segmentation with a class set similar to SemanticKITTI.} Specifically, in $T_0$ class \textit{sky} is substituted with \textit{parking} and \textit{other-ground}, in $T_1$ 
  \textit{traffic-light} and \textit{walls} with \textit{trunk}, and in $T_2$ \textit{train}, \textit{bus} and \textit{rider} are replaced with \textit{motorcyclist}, \textit{bicylist} and \textit{other-vehicle}. Such classes are \underline{underlined} in Tab.~\ref{tab:dset-subdivision}. 
  Sequence subdivisions are chosen to maximize labeled classes in each group {while keeping semantic consistency and balancing the total amount of points/acquisitions within each training subset.} Fig.~\ref{fig:chart} shows the percentage of labeled points within each group normalized for each class $c$ on the cardinality of the specific class. 

\subsubsection{Coarse-to-Fine Partition (C2F)}\label{ssec:c2f}
Fig.~\ref{fig:hierarchy} shows the second partition of SemanticKITTI, where {grouping is made by splitting labels into three learning steps, and labels are clustered to guarantee semantic consistency}. The first coarse set resembles the semantics of \textbf{CIL}: classes of each $C_k^{CIL}$ form a single coarse class in $C_0^{C2F}$. $C_1^{C2F}$ is obtained by further partitioning coarse classes into 2 mid-level classes each; finally, the fine set $C_2^{C2F} = C$, \ie, contains all the classes of SemanticKITTI. Note that this partition shares the dataset split of sequences with \textbf{CIL} setup reported in Tab~\ref{tab:dset-subdivision}.

\subsection{CL Strategies}
\label{ssec:clstrategies}

In standard training, \textit{background} points are usually masked out from the learning procedure and the model weights are optimized using the cross-entropy objective
\begin{equation}
        \mathcal{L}_{ce} = -\frac{1}{|T_k|} \sum_{{\rm \bf X}_n \in T_k}\sum_{c \in C_{k-1}} {\rm \bf Y}_n[c]\cdot \log (\mathcal{M}_{k}({\rm \bf X}_n)[c])
\end{equation}
where ${\rm \bf Y}_n[c]$ the one-hot encoded ground truth and $\mathcal{M}_{k}({\rm \bf X}_n)[c]$ is the prediction score for class $c$.
In the incremental learning setup, {we assume that at each incremental training step $k$ only samples from new classes $C_k$ are available.} In order to include this new information in the model,  weights are initialized from {previous model $\mathcal{M}_{k-1}$ ($k\geq1$)} and the new class set $C_{0\rightarrow k}$ is then learned by optimizing the standard objective $\mathcal{L}_{ce}(\mathcal{M}_{k}; C_{0\rightarrow k}, T_k)$ with data from the current training partition $T_k$. 
However, {a na\"ive fine tuning} on novel classes leads to catastrophic forgetting, and therefore, by including points from past classes (no labels) as \textit{background}, we can recover previous knowledge by self-predicting old labels. To this extent, we consider two strategies: Knowledge Distillation (KD) \cite{Li2016LearningWF} and Background Self-Inpainting \cite{maracani2021recall}. 
Fig.~\ref{fig:strategies} provides a visual summary of these techniques.

\begin{figure}[t]
    \centering
    \includegraphics[width=0.85\linewidth]{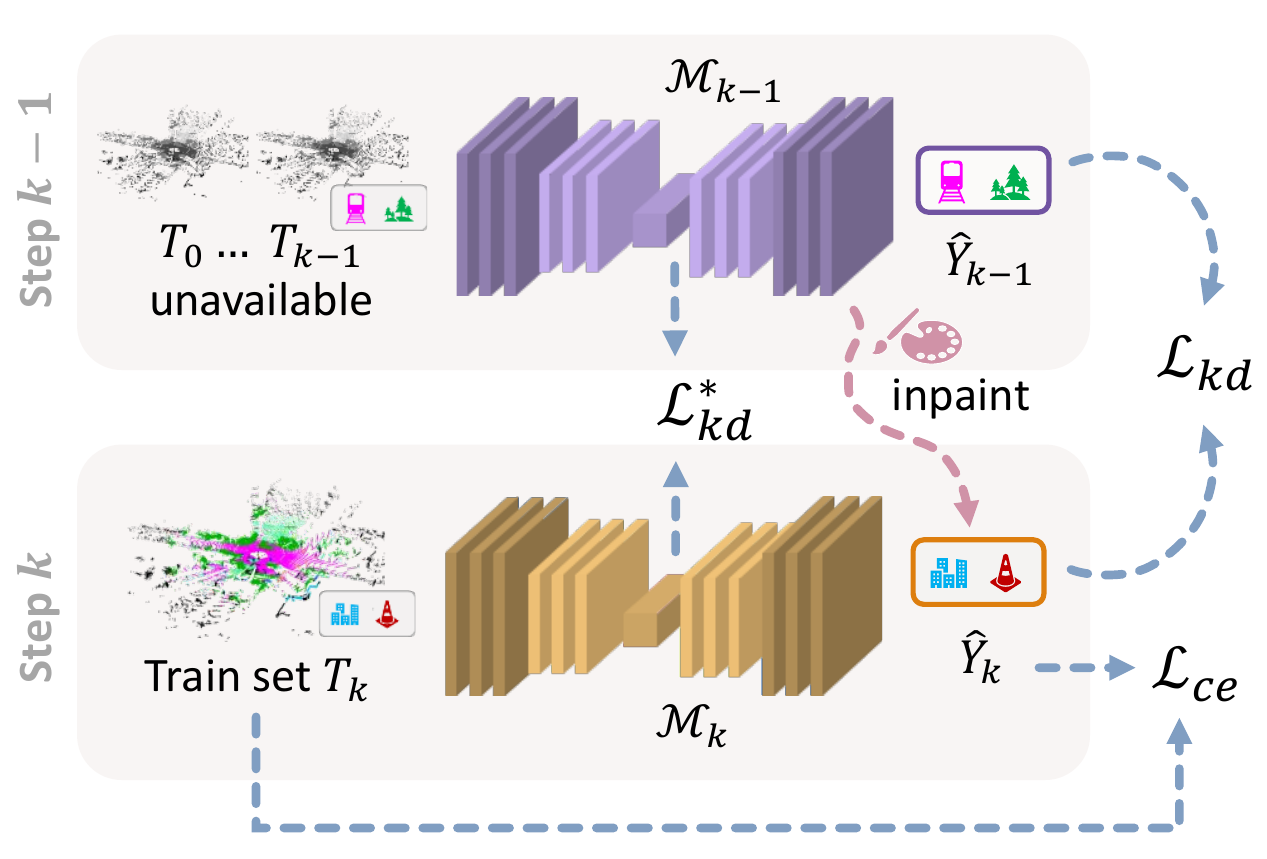}
    \caption{Summary of continual learning strategies employed. At the current learning step ($k$) only training set $T_k$ and previous model $\mathcal{M}_{k-1}$ are available.}
    \label{fig:strategies}
\end{figure}

\begin{figure}[t]
    \centering
    \includegraphics[width=0.9\linewidth]{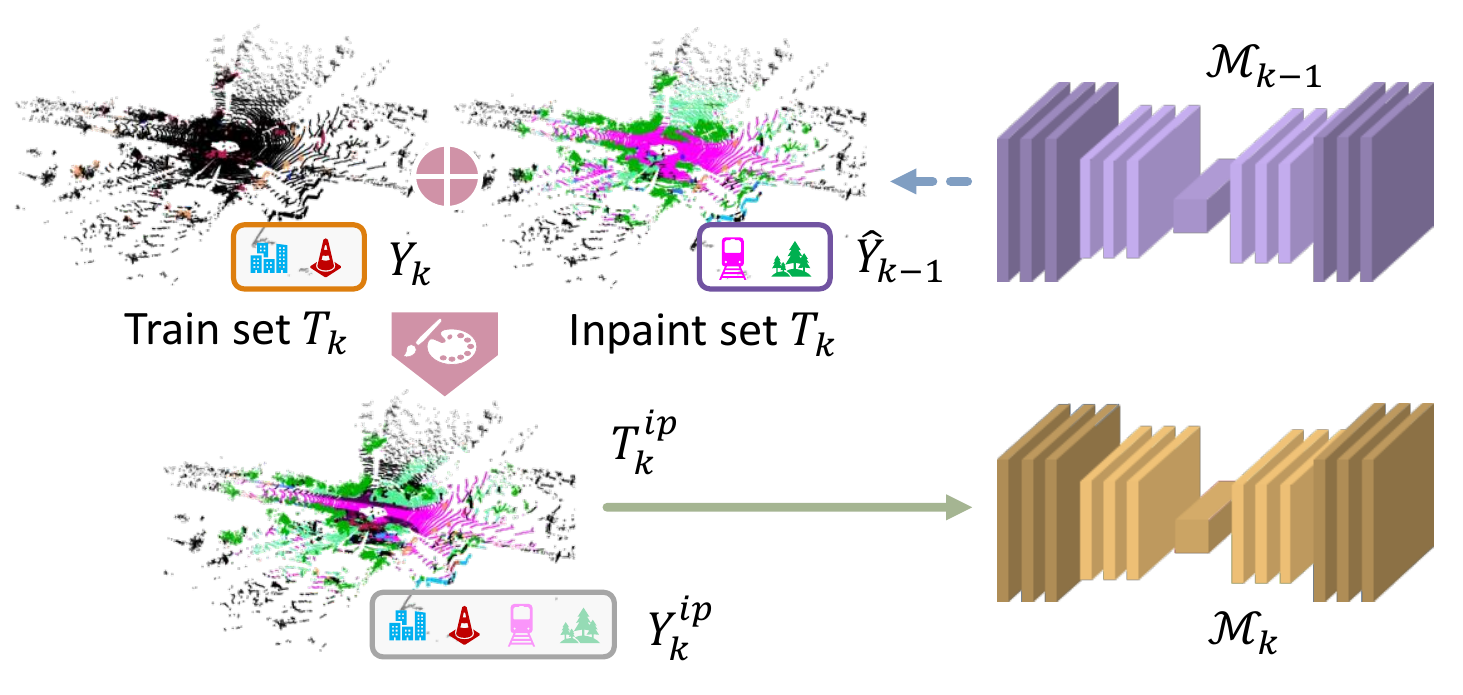}
    \caption{Background Self-Inpainting. Labels for old  classes $C_{k-1}$ are predicted by previous model $\mathcal{M}_{k-1}$ and Inpainted onto \textit{background} labels of current training set $T_k$.}
    \label{fig:inpaint}
\end{figure}

\subsubsection{Knowledge Distillation} \label{ssec:kd-o}
Knowledge Distillation (KD) is the process of transferring knowledge from a model to another (typically smaller) \cite{hinton2015KD,Hou_2022_CVPR}.
In our case, we apply knowledge distillation to recall past classes while learning new ones and to deal with the background shift phenomenon. Old labels are recovered from the previous step model $\mathcal{M}_{k-1}$.
KD is modeled as an additional objective function {$\mathcal{L}_{kd}$} tuned by a hyperparameter $\lambda$, 
obtaining as a final objective for model $\mathcal{M}_k$:
\begin{equation}
    \mathcal{L} = \mathcal{L}_{ce}(y_k,\hat{y}_k) + \lambda \cdot \mathcal{L}_{kd}(\hat{y}_{k-1},\hat{y}_{k})
\end{equation}
where $y_k$ are the ground truth labels of step $k$, $\hat{y}_{k}$ are the predictions of step $k$ and $\hat{y}_{k-1}$ are the prediction labels of step $k-1$. At step $k = 0$ KD is not applied, as at that stage we lack any prior knowledge of past classes.

\paragraph{Output Level Distillation ($\mathcal{L}_{kd}$)} is typically modeled as an additional Cross-Entropy \cite{xentropy}, where labels for past classes are obtained from previous model predictions:
\begin{equation}\small
    \mathcal{L}_{kd}=-\frac{1}{|T_k|} \sum_{{\rm \bf X}_n \in T_k}\sum_{c \in C_{k-1} \cup C_k} \mathcal{M}_{k-1}({\rm \bf X}_n)[c]  \cdot \log (\mathcal{M}_{k}({\rm \bf X}_n)[c])
\end{equation}
where $\mathcal{M}_{k}({\rm \bf X}_n)[c]$ and $\mathcal{M}_{k-1}({\rm \bf X}_n)[c]$ are the prediction scores for class $c$, obtained with current and previous models, respectively. KD can be modeled either by masking predictions of each model $\mathcal{M}_{k}$ on classes $C_k$ or joining all the predictions of unknown classes \cite{cermelli2020modeling}.
An alternative formulation updates also the previous model to refine predictions \cite{toldo2022learning}.
For Coarse-to-Fine CL, the predictions for classes within the same coarse class are summed before applying the Cross-Entropy loss. 

\begin{table*}
\begin{minipage}[t]{0.5\linewidth}
    \centering    
    \setlength{\tabcolsep}{0.25em}
    \renewcommand{\arraystretch}{0.8}
    \small
    \begin{tabular}{l|c|ccc|cccc}
    \multicolumn{9}{c}{\textbf{Cylinder3D}} \\
    \toprule
& {Step 0} & \multicolumn{3}{c}{{Step 1}} \vline &\multicolumn{4}{c}{{Step 2}}\\
Method & \rotatebox{90}{{{mIoU$_{0}$}}} &  \rotatebox{90}{{{mIoU$_{0}$}}} &  \rotatebox{90}{{{mIoU$_{1}$}}} &  \rotatebox{90}{{{mIoU$_{0,1}$}}} &  \rotatebox{90}{{{mIoU$_{0}$}}} &  \rotatebox{90}{{{mIoU$_{1}$}}} &  \rotatebox{90}{{{mIoU$_{2}$}}} &  \rotatebox{90}{{{mIoU$_{0,1,2}$}}}\\
\midrule
Baseline\dag & | & | & | & | & | & | & | & 54.1 \\
\midrule
Sequential & 55.4 & 52.7 & 46.3 & \textbf{50.1} & 52.8 & 50.6 & 32.7 & \textbf{44.9} \\ 
Sequential Masked & 55.4 & 0.0 & 77.4 & 20.4 & 0.0 & 0.0 & 25.7 & 10.8 \\ 
Disjoint & 55.4 & 0.0 & 63.3 & 20.0 & 0.0 & 0.0 & 25.2 & 10.6 \\
\midrule
{Output KD} \cite{Li2016LearningWF} & 55.4 & 45.6 & 41.1 & 43.5 & 47.4 & 32.0 & 24.5 & 33.7 \\ 
Self-Inpaint \cite{maracani2021recall} & 55.4 & 51.2 & 47.1 & \underline{49.4} & 46.2 & 36.8 & 26.9 & \underline{35.6} \\
\bottomrule
    \end{tabular}
\end{minipage} 
\hfill
\begin{minipage}[t]{0.5\linewidth}
    \centering    
    \setlength{\tabcolsep}{0.25em}
    \renewcommand{\arraystretch}{0.8}
    \small
    \begin{tabular}{l|c|ccc|cccc}
    \multicolumn{9}{c}{\textbf{RandLA-Net}} \\
    \toprule
& {Step 0} & \multicolumn{3}{c}{{Step 1}} \vline &\multicolumn{4}{c}{{Step 2}}\\
Method & \rotatebox{90}{{{mIoU$_{0}$}}} &  \rotatebox{90}{{{mIoU$_{0}$}}} &  \rotatebox{90}{{{mIoU$_{1}$}}} &  \rotatebox{90}{{{mIoU$_{0,1}$}}} &  \rotatebox{90}{{{mIoU$_{0}$}}} &  \rotatebox{90}{{{mIoU$_{1}$}}} &  \rotatebox{90}{{{mIoU$_{2}$}}} &  \rotatebox{90}{{{mIoU$_{0,1,2}$}}}\\
\midrule
Baseline\dag & | & | & | & | & | & | & | & 47.2 \\
\midrule
Sequential & 49.0 & 57.0 & 39.0 & \textbf{48.8} & 58.0 & 48.4 & 28.2 & \underline{42.9} \\ 
Sequential Masked & 49.0 & 0.0 & 48.5 & 22.0 & 0.0 & 0.0 & 31.4 & 13.2 \\ 
Disjoint & 49.0 & 0.0 & 37.9 & 18.0 & 0.0 & 0.0 & 26.1 & 11.0  \\ 
\midrule
Output KD \cite{Li2016LearningWF} & 49.0 & 56.2 & 38.0 & \underline{47.9} & 58.5 & 49.1 & 31.0 & \textbf{44.4}\\ 
Self-Inpaint \cite{maracani2021recall} & 49.0 & 56.1 & 34.9 & 46.5 & 57.5 & 43.2 & 22.6 & 39.0 \\
\bottomrule
    \end{tabular}
\end{minipage}
            \caption{Results of \textbf{Cylinder3D} and \textbf{RandLA-Net} \textbf{(CIL)}. \textbf{Bold} denotes the best result, \underline{underline} the second best. \dag: baselines retrained with our configuration.}\label{tab:Cyl-Rand}
\end{table*}

\begin{table*}[t]
\centering \small
    \setlength{\tabcolsep}{0.25em}
    \renewcommand{\arraystretch}{0.8}
\begin{tabular}{c|ccccccccccccccccccc|ccc}
\toprule
 \textbf{method} & \rotatebox{90}{car} & \rotatebox{90}{bicycle} & \rotatebox{90}{motorcycle} & \rotatebox{90}{truck} & \rotatebox{90}{other-vehicle} & \rotatebox{90}{person} & \rotatebox{90}{bicyclist} & \rotatebox{90}{motorcyclist} & \rotatebox{90}{road} & \rotatebox{90}{parking} & \rotatebox{90}{sidewalk} & {\rotatebox{90}{other-ground}} & \rotatebox{90}{building} & \rotatebox{90}{fence} & \rotatebox{90}{vegetation} & \rotatebox{90}{trunk} & \rotatebox{90}{terrain} & \rotatebox{90}{pole} & \rotatebox{90}{traffic-sign} & \rotatebox{90}{ \bf mIoU $\uparrow$} & \rotatebox{90}{ \bf Std Dev. $\sigma \downarrow$} \\ \midrule
 Baseline\dag & 90.7 & 1.6 & 14.6 & 53.9 & 37.1 & 30.5 & 54.2 & 0.0 & 90.9 & 35.4 & 75.1 & 2.2 & 84.0 & 46.3 & 84.9 & 47.1 & 69.7 & 51.7 & 27.6 & 47.2 & 30.1 \\
 \midrule
 Sequential & 89.3 & 0.0 & 12.5 & 59.4 & 33.0 & 8.2 & 23.2 & 0.0 & 90.5 & 26.0 & 73.9 & 0.8 & 84.2 & 47.8 & 85.1 & 51.9 & 71.6 & 49.3 & 8.7 & \underline{42.9} & \underline{33.0} \\
Output KD \cite{Li2016LearningWF} & 88.6 & 0.0 & 11.6 & 52.6 & 32.2 & 10.2 & 52.5 & 0.0 & 90.4 & 28.5 & 73.9 & 0.0 & 84.6 & 47.0 & 85.0 & 55.9 & 73.3 & 51.4 & 6.7 & \textbf{44.4} & \textbf{32.8} \\ 
Output UKD \cite{cermelli2020modeling} & 88.0 & 0.0 & 7.3 & 27.2 & 17.8 & 0.9 & 16.9 & 0.0 & 90.3 & 29.8 & 74.3 & 2.5 & 83.3 & 47.4 & 84.8 & 50.8 & 72.8 & 44.0 & 1.3 & 38.9 & 34.3 \\ 
Output XKD \cite{toldo2022learning} & 86.5 & 0.0 & 11.6 & 59.3 & 21.0 & 2.5 & 45.2 & 0.0 & 90.8 & 32.3 & 75.1 & 0.2 & 83.0 & 44.3 & 85.0 & 46.0 & 72.6 & 46.5 & 8.9 & 42.7 & \underline{33.0}\\ 
Feature KD (L$_2$) & 88.4 & 0.0 & 13.2 & 57.0 & 16.4 & 9.0 & 35.1 & 0.0 & 90.3 & 22.4 & 73.9 & 0.5 & 82.5 & 44.1 & 84.8 & 52.6 & 72.8 & 49.3 & 0.1 & 41.7 & 33.6\\
Feature KD (L$_1$) & 87.5 & 0.0 & 10.9 & 36.1 & 35.9 & 0.0 & 39.9 & 0.0 & 90.4 & 32.1 & 74.2 & 0.1 & 82.4 & 40.7 & 85.7 & 53.4 & 75.2 & 46.6 & 1.1 & 41.7 & 33.4 \\
Both KD \cite{michieli2019} & 85.3 & 0.0 & 10.2 & 42.9 & 20.2 & 0.0 & 33.9 & 0.0 & 90.0 & 24.8 & 72.9 & 0.4 & 82.7 & 42.8 & 84.1 & 43.7 & 72.0 & 44.5 & 0.0 & 39.5 & 33.3\\
Self-Inpaint \cite{maracani2021recall} & 86.3 & 0.0 & 7.3 & 54.5 & 5.3 & 0.1 & 27.0 & 0.0 & 90.4 & 25.0 & 74.1 & 1.6 & 82.9 & 42.3 & 82.7 & 45.3 & 71.1 & 45.5 & 0.0 & 39.0 & 34.4 \\
\bottomrule
    \end{tabular}
            \caption{Per-class mIoU results of \textbf{RandLA-Net} \textbf{(CIL)}. \textbf{Bold} denotes the best result, \underline{underline} the second best. \dag: baseline retrained with our configuration.}\label{tab:perclass-CIL}
\end{table*}

\paragraph{Feature Level Distillation ($\mathcal{L}^{*}_{kd}$)} is typically modeled as L$_p$ (L$_2$ or L$_1$) norm between the features of current and previous models:
\begin{equation}
    \mathcal{L}^{*}_{kd}=\frac{|| \mathcal{E}_{k-1}({\rm \bf X}_n) - \mathcal{E}_{k}({\rm \bf X}_n)||_{p}}{|T_k|}
\end{equation}
where $\mathcal{E}_{k}({\rm \bf X}_n)[c]$ and $\mathcal{E}_{k-1}({\rm \bf X}_n)[c]$ are the features of current and previous models, respectively. The choice of this design is motivated by the fact that this strategy provides learning of a feature map rather than a classification map and distance loss provides an alignment of such embeddings. KD at feature and output levels can be also combined to obtain joint effects from both.

\subsubsection{Background Self-Inpainting}\label{ssec:bg-ip}
Background Self-Inpainting is a simple self-teaching mechanism that provides pseudo labeling of current \textit{background} samples \cite{maracani2021recall}, reducing the background shift while bringing a regularization effect similar to knowledge distillation. At every step $k$ with training set $T_k$, we take the \textit{background} points of each ground truth map $Y_k$ and we label it with the associated prediction from the previous model $\mathcal{M}_{k-1}$ (Fig.~\ref{fig:inpaint}). More formally, we replace each original label map $Y_k$ available at step $k > 0$ with its inpainted version $Y_{k}^{ip}$:
\begin{equation}
   {Y}_k^{ip}[i] = \begin{cases} {Y_k[i]} & \text {if} \quad {Y} [i] \in {C}_k \\ 
   {\rho \cdot \hat{Y}_{k-1}[i]} & \text {otherwise} \end{cases}  
\end{equation}
where $(X_k, Y_k) \in T_k$. $[i]$ denotes $i$-th sample in $T_k$ and $\rho$ defines the inpaining rule. $\rho$ is designed as:
\begin{equation}
\rho = \begin{cases} 1 & \text{if} \quad \hat{\mathcal{Y}}^{*}_k[i] - \hat{\mathcal{Y}}^{**}_k[i] > \tau_1 \quad \text{and} \quad \hat{\mathcal{Y}}^{*}_k[i] > \tau_2\\
0 & \quad \text{otherwise}. \end{cases}
\end{equation}
$\hat{\mathcal{Y}}^{*}_k[i]$ and $\hat{\mathcal{Y}}^{**}_k[i]$ denote the sets of first and second maximum in the softmax predictions $\hat{\mathcal{Y}}_k$ at step $k$.
Labels at step $k = 0$ are not inpainted, as at that stage we lack any prior knowledge of past classes. When background inpainting is performed, each set $T^{ip}_k \subset X \times Y_{C_{0\rightarrow k}}$ ($k > 0$) contains all samples of $T_k$ after being inpainted.

\section{Experimental Results}\label{sec:exp}

For the experimental evaluation, we use RandLA-Net \cite{hu2019randla} 
(a widely used lightweight point based architecture)
and compare results with Cylinder3D \cite{zhu2021cylindrical,zhu2021cylindrical-tpami} (state-of-the-art model based on voxels and point-wise prediction refinements). Note that the approach is generalizable to any other segmentation architecture for point clouds.
We used SemanticKITTI \cite{behley2019iccv,geiger2012cvpr} as a reference dataset, with classes and data partitioned into learning steps according to \textbf{CIL} and \textbf{C2F} splits, described in Sec.~\ref{sec:dataset}.
The model performance is measured by the mean Intersection over Union (mIoU) \cite{7298965}. From now on, we use the notation mIoU$_k$ to denote the mIoU computed on class set $C_k$.

We train both RandLA-Net and Cylinder3D using Adam optimizer with their standard learning rate policy, momentum, and weight decay. The initial learning rates are set respectively to $0.01$ and $0.001$ for the two models.
The incremental training has been performed decreasing the learning rate according to a polynomial decay rule with power $0.95$. Note that at step $k$ the learning rate is initialized to the last learning rate of step $k-1$ to better preserve previous weights. We set batch size $3$ and train each learning step $k$ for a number of epochs $2 \times |C_k|$, proportional to the number of new classes to be learned. 
We used PyTorch \cite{NEURIPS2019_9015} for the implementation and train both models on NVIDIA GeForce RTX 3090Ti graphics processing unit with CUDA 11.6. The training of each learning step took around 5 hours with RandLA-Net and 30 hours with Cylinder3D. The code is available at: \texttt{\href{https://github.com/LTTM/CL-PCSS}{https://github.com/LTTM/CL-PCSS}}.
In the following sections, we discuss the results obtained with \textbf{CIL} and \textbf{C2F} partitions. Baseline results are obtained by training the model from scratch on the whole dataset with all the classes, using the aforementioned configuration. 

\subsection{CIL}

We performed our first set of experiments following \textbf{CIL} configuration. First, we developed standard fine-tuning methods, knowledge distillation, and self-inpainting to compare Cylinder3D with RandLA-Net. Results are reported in Tab.~\ref{tab:Cyl-Rand}.
In general, {the obtained performances} reflect the behavior of the corresponding image-based setups. As expected, the disjoint scenario performs the worst in all the learning steps, strongly suffering the background shift phenomenon. Sequential masked improves a little the results besides showing again the effect of catastrophic forgetting when accounting for new classes. 

\begin{figure}[t]
    \centering
    \includegraphics[width=\linewidth]{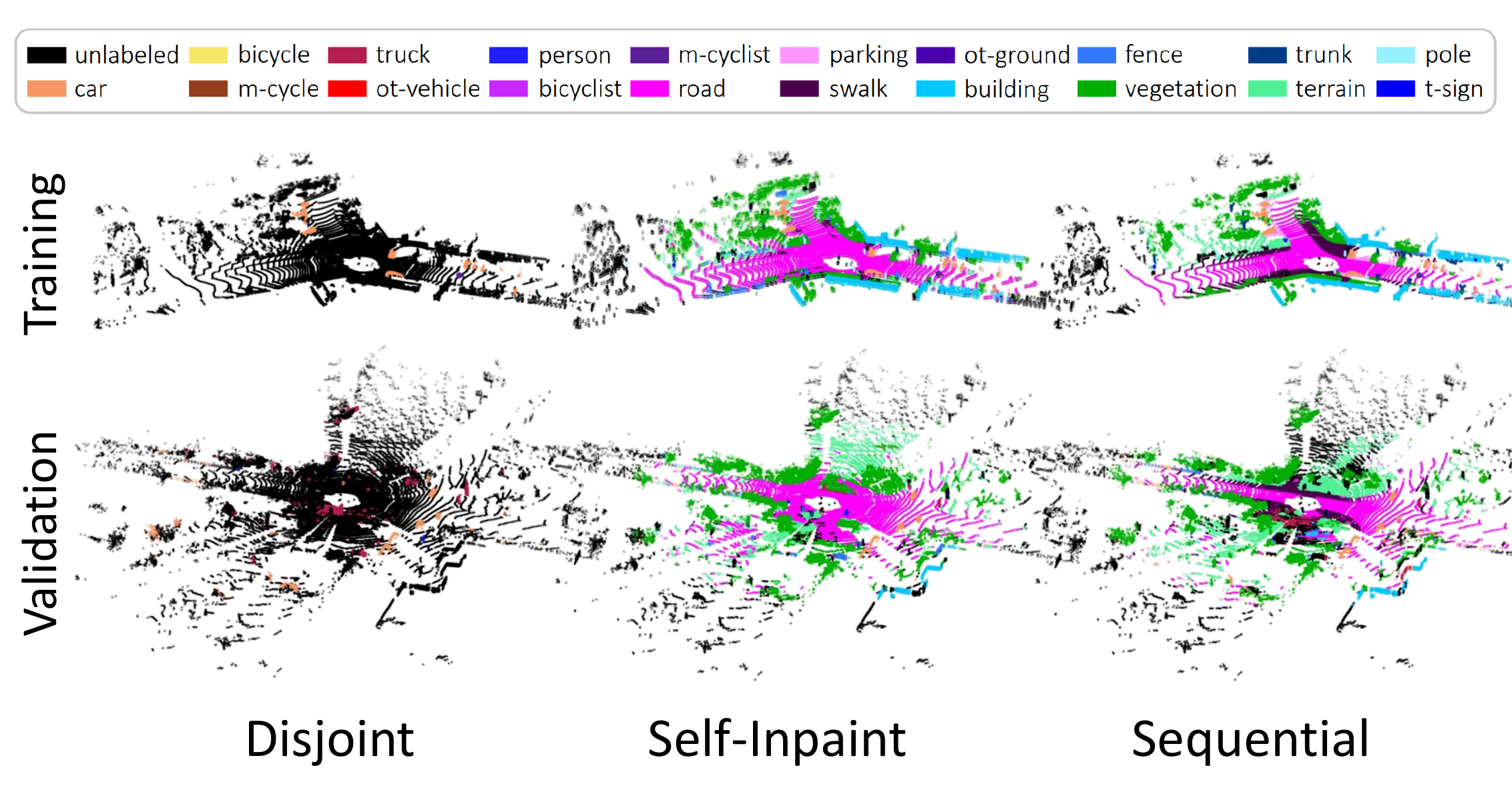} \vskip -2ex
    \caption{Some qualitative results on \textbf{CIL} with \textbf{Cylinder3D}.}
    \label{fig:qual}
\end{figure}

Overall, {Cylinder3D shows worse performance with respect to RandLA-Net within a CL framework, despite its better performance in the non-incremental scenario}. Its results on the baseline show a strong performance decrease losing $9.8\%$ mIoU in the sequential setup (best case for fine-tuning). Besides, the size of the architecture and the processing of voxels make the training time 6 times longer.
KD partially recovers knowledge from previous classes but keeps stability when learning new ones. Instead, Self-Inpainting is more performing and accounts better for new classes. {Note that, mIoU$_0$ is efficiently recovered by both methods across the steps, while mIoU$_1$ is hardly restored.} This is partially due to the fact that labels for step $k = 0$ are robust classes with many points and a general method for catastrophic forgetting mitigation can be sufficient for recovering knowledge. Contrarily, step $k = 1$ has a huge decrease in the number of labeled points that reflects the increasing difficulty in obtaining good recovery from previous training.
Self-Inpainting hardly pseudo-labels old classes, not preserving the uncertainty of predictions, and penalizing the learning of new classes. The introduction of inpainting rules (Tab.~\ref{tab:res:semKITTI-ip}) brings a slight improvement, avoiding strongly wrong predictions to be considered as true labels in the subsequent step. In this way, previous knowledge is hardly restored only in case of accurate predictions and blended into the current model in a stronger way with respect to KD. Finally, both methods perform worse with respect to the sequential setup.
Qualitative results of the self-inpainting strategy are shown in Fig.~\ref{fig:qual}. 

\begin{table}[t]
    \centering    
    \setlength{\tabcolsep}{0.2em}
    \renewcommand{\arraystretch}{0.8}
    \small
    \begin{tabular}{l|c|ccc|cccc}
    \toprule
& {Step 0} & \multicolumn{3}{c}{{Step 1}} \vline &\multicolumn{4}{c}{{Step 2}}\\
Method & \rotatebox{90}{{{mIoU$_{0}$}}} &  \rotatebox{90}{{{mIoU$_{0}$}}} &  \rotatebox{90}{{{mIoU$_{1}$}}} &  \rotatebox{90}{{{mIoU$_{0,1}$}}} &  \rotatebox{90}{{{mIoU$_{0}$}}} &  \rotatebox{90}{{{mIoU$_{1}$}}} &  \rotatebox{90}{{{mIoU$_{2}$}}} &  \rotatebox{90}{{{mIoU$_{0,1,2}$}}}\\
\midrule
Baseline\dag & | & | & | & |  & | & | & | & 47.2 \\
\midrule
Overlapped & 60.2 & 0.0 & 51.6 & \textbf{23.4} & 0.0 & 0.0 & 32.2 & \textbf{13.6}\\ 
Disjoint & 49.0 & 0.0 & 37.9 & 18.0 & 0.0 & 0.0 & 26.1 & 11.0 \\ 
\midrule
Output KD \cite{Li2016LearningWF} & 49.0 & 56.2 & 38.0 & \underline{47.9} & 58.5 & 49.1 & 31.0 & \textbf{44.4}\\ 
Output UKD \cite{cermelli2020modeling} & 49.0 & 56.1 & 35.3 & 46.6 & 59.1 & 45.4 & 19.8 & 38.9 \\ 
Output XKD \cite{toldo2022learning} & 49.0 & 56.0 & 35.4 & 46.7 & 59.3 & 45.7 & 28.3 & \underline{42.7} \\ 
Feature KD (L$_2$) & 49.0 & 55.9 & 40.7 & \textbf{49.0} & 57.5 & 45.7 & 27.4 & 41.7 \\
Feature KD (L$_1$) & 49.0 & 55.5 & 36.4 & 46.8 & 59.6 & 44.8 & 26.3 & 41.7\\
Both KD \cite{michieli2019} & 49.0 & 56.5 & 33.9 & 46.2 & 57.4 & 42.7 & 24.1 & 39.5 \\
\bottomrule
    \end{tabular}
            \caption{\textbf{RandLA-Net} \textbf{(CIL)}. \textbf{Bold} denotes the best result, \underline{underline} the second best. \dag: baseline retrained with our configuration.}\label{tab:CIL}
\end{table}

\begin{table}[t]
    \centering    \small
    \setlength{\tabcolsep}{0.4em}
    \renewcommand{\arraystretch}{0.8}
        \begin{tabular}{cc|c|ccc|cccc}
    \toprule
&  & \multicolumn{1}{c}{Step 0} \vline & \multicolumn{3}{c}{Step 1} \vline &\multicolumn{4}{c}{Step 2}\\
$\boldsymbol{\tau_1}$ & $\boldsymbol{\tau_2}$ & \rotatebox{90}{{{mIoU$_{0}$}}} &  \rotatebox{90}{{{mIoU$_{0}$}}} &  \rotatebox{90}{{{mIoU$_{1}$}}} &  \rotatebox{90}{{{mIoU$_{0,1}$}}} &  \rotatebox{90}{{{mIoU$_{0}$}}} &  \rotatebox{90}{{{mIoU$_{1}$}}} &  \rotatebox{90}{{{mIoU$_{2}$}}} &  \rotatebox{90}{{{mIoU$_{0,1,2}$}}}\\
\midrule
 \textbf{0} & \textbf{0} & 55.4 & 50.7 & 39.6 & 45.8 & 47.7 & 42.8 & 19.4 & 34.5 \\
\textbf{0} & \textbf{0.7} & 55.4 & 48.4 & 37.9 & 43.6 & 44.0 & 38.7 & 25.0 & 34.6 \\ 
\textbf{0.2} & \textbf{0} & 55.4 & 55.8 & 40.5 & 48.9 & 45.2 & 35.9 & 28.0 & 35.5 \\
\textbf{0.2} & \textbf{0.7} & 55.4 & 51.2 & 47.1 & \textbf{49.4} & 46.2 & 36.8 & 26.9 & \textbf{35.6} \\
\bottomrule
    \end{tabular} \caption{Ablation study on Self-Inpainting with \textbf{Cylinder3D}. $\tau_1, \tau_2$ are the parameters for inpainting. \textbf{Bold} denotes the best result.}
    \label{tab:res:semKITTI-ip}
\end{table}

\begin{table}[t]
    \centering    
    \setlength{\tabcolsep}{0.25em}
    \renewcommand{\arraystretch}{0.8}
    \small
    \begin{tabular}{l|c|cc|ccc}
    \toprule
& {Step 0} & \multicolumn{2}{c}{{Step 1}} \vline &\multicolumn{3}{c}{{Step 2}}\\
&  &  &\\
Method & \rotatebox{90}{{{mIoU$_{0}$}}} &  \rotatebox{90}{{{mIoU$_{0}$}}} &  \rotatebox{90}{{{mIoU$_{1}$}}} &  \rotatebox{90}{{{mIoU$_{0}$}}} &  \rotatebox{90}{{{mIoU$_{1}$}}} &  \rotatebox{90}{{{mIoU$_{2}$}}} \\
\midrule
Baseline\dag & | & | & | & |  & | & 47.2 \\
\midrule
Fine-Tuning & 86.7 & 74.6 & \textbf{74.6} & 51.0 & 51.0 & \textbf{47.1}\\ 
Output KD & 86.7 & 52.9 & 52.9 & 48.8 & 48.8 & 45.1 \\ 
Feature KD (L$_1$) & 86.7 & 57.2 & 57.2 & 47.1 & 47.1 & 42.8 \\
\bottomrule
    \end{tabular}
            \caption{\textbf{RandLA-Net} on \textbf{SemanticKITTI (C2F)}. \textbf{Bold} denotes the best result. \dag: baseline retrained with our configuration.}\label{tab:C2F}
\end{table}

\begin{table*}[t]
\centering \small
    \setlength{\tabcolsep}{0.25em}
    \renewcommand{\arraystretch}{0.8}
\begin{tabular}{c|ccccccccccccccccccc|ccc}
\toprule
 \textbf{method} & \rotatebox{90}{car} & \rotatebox{90}{bicycle} & \rotatebox{90}{motorcycle} & \rotatebox{90}{truck} & \rotatebox{90}{other-vehicle} & \rotatebox{90}{person} & \rotatebox{90}{bicyclist} & \rotatebox{90}{motorcyclist} & \rotatebox{90}{road} & \rotatebox{90}{parking} & \rotatebox{90}{sidewalk} & {\rotatebox{90}{other-ground}} & \rotatebox{90}{building} & \rotatebox{90}{fence} & \rotatebox{90}{vegetation} & \rotatebox{90}{trunk} & \rotatebox{90}{terrain} & \rotatebox{90}{pole} & \rotatebox{90}{traffic-sign} & \rotatebox{90}{\textbf{mIoU $\uparrow$}} & \rotatebox{90}{\textbf{Std Dev. $\sigma \downarrow$}} \\ \midrule
  Baseline\dag & 90.7 & 1.6 & 14.6 & 53.9 & 37.1 & 30.5 & 54.2 & 0.0 & 90.9 & 35.4 & 75.1 & 2.2 & 84.0 & 46.3 & 84.9 & 47.1 & 69.7 & 51.7 & 27.6 & 47.2 & 30.1 \\ \midrule
 Fine-Tuning & 89.7 & 6.2 & 17.4 & 47.0 & 24.8 & 28.3 & 65.5 & 0.0 & 88.8 & 30.7 & 73.0 & 0.6 & 83.6 & 45.8 & 87.2 & 55.4 & 75.1 & 50.8 & 24.4 & \textbf{47.1} & \textbf{30.7} \\

Output KD & 90.3 & 0.2 & 16.6 & 53.2 & 30.0 & 14.8 & 42.1 & 0.0 & 90.7 & 26.9 & 74.1 & 0.1 & 84.5 & 43.9 & 86.5 & 56.8 & 73.9 & 52.0 & 21.1 & 45.1 & 31.9 \\ 
Feature KD (L$_1$) & 89.2 & 0.0 & 12.5 & 65.7 & 29.6 & 3.6 & 41.4 & 0.0 & 91.2 & 37.9 & 74.3 & 0.8 & 84.8 & 45.4 & 86.2 & 50.1 & 72.9 & 48.7 & 18.2 & 44.9 & 32.7\\
\bottomrule
    \end{tabular}
            \caption{Per-class mIoU results on \textbf{C2F} with \textbf{RandLA-Net}. \textbf{Bold} denotes the best result. \dag: baseline retrained with our configuration.} \label{tab:perclass-C2F}
\end{table*}

\newcolumntype{a}{>{\columncolor{AntiqueWhite1}}c}
\newcolumntype{b}{>{\columncolor{Honeydew1}}c}
\newcolumntype{d}{>{\columncolor{LemonChiffon1}}c}

\begin{table*}[t]
\centering \small
    \setlength{\tabcolsep}{0.25em}
    \renewcommand{\arraystretch}{0.8}
\begin{tabular}{c|aaaaa|aaa|bbbb|bb|dd|ddd|cc|cc}
\toprule
 \textbf{step} & \rotatebox{90}{car} & \rotatebox{90}{bicycle} & \rotatebox{90}{motorcycle} & \rotatebox{90}{truck} & \multicolumn{1}{a}{\rotatebox{90}{other-vehicle}} & \rotatebox{90}{person} & \rotatebox{90}{bicyclist} & \multicolumn{1}{a}{\rotatebox{90}{motorcyclist}} & \rotatebox{90}{road} & \rotatebox{90}{parking} & \rotatebox{90}{sidewalk} & \multicolumn{1}{b}{\rotatebox{90}{other-ground}} & \rotatebox{90}{vegetation} & \multicolumn{1}{b}{\rotatebox{90}{terrain}} & \rotatebox{90}{building} & \multicolumn{1}{d}{\rotatebox{90}{fence}} & \rotatebox{90}{trunk} & \rotatebox{90}{pole} & \rotatebox{90}{traffic-sign} & \rotatebox{90}{\textbf{mIoU $\uparrow$} \cite{7298965}} & \rotatebox{90}{\textbf{Std Dev. $\sigma \downarrow$}} & \rotatebox{90}{\textbf{PA $\uparrow$} \cite{JohnsonRoberson2016DrivingIT}} & \rotatebox{90}{\textbf{PP $\uparrow$} \cite{pointpillars}} \\ \midrule
\textbf{0} & \multicolumn{8}{a}{85.1} \vline & \multicolumn{6}{b}{95.6} \vline & \multicolumn{5}{d}{79.5} \vline & 86.7 & 8.2 & 93.0 & 92.6\\
\midrule
\textbf{1} & \multicolumn{5}{a}{89.1} \vline & \multicolumn{3}{a}{43.4} \vline & \multicolumn{4}{b}{92.2} \vline & \multicolumn{2}{b}{89.8} \vline & \multicolumn{2}{d}{81.7} \vline & \multicolumn{3}{d}{51.3} \vline & 74.6 & 21.5 & 80.4 & 89.6\\
\midrule
\textbf{2} & 89.7 & 6.2 & 17.4 & 47.0 & 24.8 & 28.3 & 65.5 & 0.0 & 88.8 & 30.7 & 73.0 & 0.6 & 87.2 & 75.1 & 83.6 & 45.8 & 55.4 & 50.8 & 24.4 & 47.1 & 30.7 & 56.2 & 67.7 \\
\bottomrule
    \end{tabular}
            \caption{Per-class Coarse-to-Fine results, standard fine-tuning. Different colors are associated with different Coarse partitions. } \label{tab:perstep-C2F} \vskip -2ex
\end{table*}

On the other hand, RandLA-Net shows great suitability for continual learning settings, with a decrease of only $4.7\%$ mIoU in the sequential setup, with respect to the baseline. Surprisingly, recovering knowledge via distillation obtains $44.4\%$ mIoU, improving the sequential setup of $1.5\%$ mIoU with a gap from the baseline of only $2.8\%$ mIoU. This is again a consequence of the label frequency of each split: early steps have a large number of labeled samples, while later steps contain only a few labeled samples. The network is subject to this imbalance when it is trained only with $\mathcal{L}_{ce}$, both in sequential Fine-Tuning and Self-Inpainting.

For this extent, we led extensive experiments with RandLA-Net on Knowledge Distillation, considering both the case of Distilling output predictions and intermediate features.
Tab.~\ref{tab:CIL} reports the overall mIoU results across learning steps, while Tab.~\ref{tab:perclass-CIL} reports per-class results in the final learning step.
Both distillation methods consistently improve the overall results. The overall best solution  is represented by the standard output level distillation, but also KD at the feature level performs quite well.
However, in general, per-class results show that the incremental setup emphasizes the gap in performance between frequent and infrequent classes. For example, class \textit{car} obtains always good results around $90\%$ mIoU, while class \textit{bicycle} (belonging to the same partition) obtains $1.6\%$ mIoU on the baseline but is never classified correctly in the incremental setup. Similarly, \textit{traffic-sign} obtains $27.6\%$ mIoU in the baseline but its performance decrease drastically when incremental learning is applied; in Self-inpainting it is never classified correctly.
Indeed, the lowest standard deviation is achieved by the baseline model.

Finally, worth to be mentioned is the result of the Overlapped setup (reported in Tab.~\ref{tab:CIL}) in comparison with the Disjoint one: training on the whole dataset brings as expected an improvement ($2.6\%$ on the final mIoU) but still suffers for catastrophic forgetting.

\subsection{Coarse-to-Fine}

Results on the Coarse-to-Fine partition are reported in Tab.~\ref{tab:perclass-C2F}. Fine-tuning on coarse classes obtains results comparable with the standard baseline training ($47.1\%$ vs $47.2\%$ mIoU), despite a higher standard deviation among classes. Note that this tuning is performed on the split sequences of Tab.~\ref{tab:dset-subdivision}, while general Coarse-to-Fine approaches rely on the Overlapped approach. Contrarily to \textbf{CIL}, a direct application of Knowledge Distillation leads here to poor performance, both when applied at the output level and intermediate feature space, losing $2.0\%$ and {$2.2\%$} mIoU respectively. Hence, this setting requires the application of more sophisticated loss functions in order to obtain a substantial improvement.
Tab.~\ref{tab:perstep-C2F} reports the results of the \textbf{C2F} fine-tuning in terms of mIoU, Point Accuracy (PA) \cite{JohnsonRoberson2016DrivingIT} and Point Precision (PP) \cite{pointpillars}.
Results in the first step achieve higher accuracy with a low standard deviation, which increases with learning steps (classes are fairly distributed up to step $k=1$). This result suggests that additional intermediate steps could be beneficial to improve the final performance, which in general requires further refinement.

\section{Conclusion}\label{sec:conclusion}

The paper tackled the problem of class incremental continual learning for LiDAR semantic segmentation.
We formally partitioned SemanticKITTI into semantically-consistent groups, and we evaluated partitions on CL strategies addressing different techniques to prevent catastrophic forgetting. 
A comparison of RandLA-Net and Cylinder3D performance shows that the former (point based and lightweight) fits better into the class-incremental setup.
Experiments on our proposed subdivision of SemanticKITTI prove the efficiency of CL strategies in alleviating catastrophic forgetting, even on sparse data.
The overall problem still requires improvements as the performance is lower than the ones achieved after standard single-step training. 
Future work will expand the experimental framework by introducing class balancing constraints among previous and current experiences, and geometric constraints, designed ad hoc for LiDAR point clouds.


\section*{Acknowledgements}
This work was partially supported by the European Union under the Italian National Recovery and Resilience Plan (NRRP) of NextGenerationEU, a partnership on “Telecommunications of the Future” (PE0000001 - program “RESTART”).

{\small
\bibliographystyle{ieee_fullname}

}

\end{document}